\documentclass[sigconf,nonacm]{acmart}
\usepackage{CJKutf8} 
\usepackage{xcolor}
\usepackage{subcaption}
\usepackage{graphicx}

\AtBeginDocument{%
  }

\begin{document}
\begin{CJK}{UTF8}{gbsn} 

\title{Model Inversion in Split Learning for Personalized LLMs: New Insights from Information Bottleneck Theory}

\author{Yunmeng Shu}
\email{fuyi_verty@sjtu.edu.cn}
\affiliation{%
  \institution{Shanghai Jiao Tong University}
  \city{Shanghai}
  \country{China}
}

\author{Shaofeng Li}
\email{
shaofengli@seu.edu.cn}
\affiliation{%
  \institution{Southeast University}
  \city{Nanjing}
  \country{China}
}

\author{Tian Dong}
\email{tian.dong@sjtu.edu.cn}
\affiliation{%
  \institution{Shanghai Jiao Tong University}
  \city{Shanghai}
  \country{China}
}

\author{Yan Meng}
\email{yan_meng@sjtu.edu.cn}
\affiliation{%
  \institution{Shanghai Jiao Tong University}
  \city{Shanghai}
  \country{China}
}

\author{Haojin Zhu}
\email{zhu-hj@cs.sjtu.edu.cn}
\affiliation{%
  \institution{Shanghai Jiao Tong University}
  \city{Shanghai}
  \country{China}
}

\begin{abstract}

Personalized Large Language Models (LLMs) have become increasingly prevalent, showcasing the impressive capabilities of models like GPT-4. This trend has also catalyzed extensive research on deploying LLMs on mobile devices. Feasible approaches for such edge-cloud deployment include using split learning. However, previous research has largely overlooked the privacy leakage associated with intermediate representations transmitted from devices to servers. 
This work is the first to identify model inversion attacks in the split learning framework for LLMs, emphasizing the necessity of secure defense. 
For the first time, we introduce mutual information entropy to understand the information propagation of Transformer-based LLMs and assess privacy attack performance for LLM blocks. To address the issue of representations being sparser and containing less information than embeddings, we propose a two-stage attack system in which the first part projects representations into the embedding space, and the second part uses a generative model to recover text from these embeddings. This design breaks down the complexity and achieves attack scores of 38\%-75\% in various scenarios, with an over 60\% improvement over the SOTA. 
This work comprehensively highlights the potential privacy risks during the deployment of personalized LLMs on the edge side. 
\end{abstract}




\maketitle

\section{Introduction}

In the rapidly evolving landscape of Artificial Intelligence (AI), particularly with the advancements in Large Language Models (LLMs) like GPT-4 \cite{openai2024gpt4technicalreport}, there has been a significant shift towards enhancing personalization through extensive training data and sophisticated algorithms. Initiatives such as OpenAI's Agent program and advancements in Retrieval-Augmented Generation (RAG)\cite{lewisRetrievalAugmentedGenerationKnowledgeIntensive2021} technology exemplify this trend, aiming to tailor LLMs with individual data to improve accuracy and relevance, thereby underscoring their growing importance in both personal and professional domains. 

\begin{figure}
\centering
\includegraphics[width=0.5\textwidth,height=0.3\textwidth]{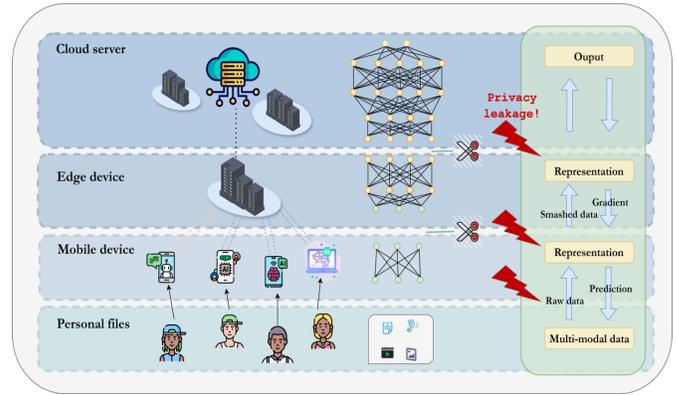}
\caption{Collaborative training in split learning scenario} 
\label{joint training system}
\end{figure}

However, leveraging LLMs through traditional cloud computing methods introduces several challenges, such as latency, bandwidth resources, and privacy concerns, especially given the multi-modal nature of modern AI applications. To bridge the gap between data locality and model scalability for personalized LLM training, split learning emerges as a solution for collaborative training of deep networks. As depicted in Fig. \ref{joint training system}, users can segment the model at chosen split points and engage in collaborative training. Recent studies on split LLMs show that distributing multi-head self-attention layers across multiple devices enables joint parallel computation. 

However, the security and privacy of split personalized LLMs have yet to receive sufficient attention. Some claim that when using edge-cloud collaborative method, a curious-but-honest server no longer accesses raw data, which ostensibly improves security. But do these methods truly protect local users' privacy? Intermediate representations contain substantial information that, if inverted, can lead to severe privacy breaches. Attackers can exploit intermediate representations of input data for privacy inversion attacks, including ``cut layer" information, ``smashed data" in computer vision, or ``hidden state" information in the field of NLP. In computer vision, inversion techniques have shown the ability to recover input images from final layer activations, while the literature on black-box attacks indicates that segmentation layer activation values can be exploited. 

In the field of natural language, there remains a noticeable gap in research addressing the risk of inversion attacks on intermediate representations in LLMs. Most research focuses only on embedding layer inversion and does not consider all intermediate representations comprehensively. Table \ref{tab:example} illustrates the potential for inversion using our approach, \textit{RevertLM}.

\begin{table}[h!]
  \caption{Examples of text inversion with RevertLM at the 20$^{\text{th}}$ decoder LLM blocks}
  \centering
  \resizebox{0.5\textwidth}{!}{%
    \begin{tabular}{p{3cm}|p{3cm}}
      \hline
      \textbf{Raw text} & \textbf{Inversed text} \\
      \hline
      \textit{\textcolor{red}{no i just make boats on the weekend . what else do you do ?}} & \textit{\textcolor{red}{no i just make boats on the weekend . what else do you do ?}} \\
      \hline
      \textit{\textcolor{red}{deep sea or} fresh water ?} & \textit{water borne  \textcolor{red}{or deep sea}?} \\
      \hline
      \textit{\textcolor{red}{what is your favorite} holiday ? mine is \textcolor{red}{christmas} .} & \textit{\textcolor{red}{which is your favorite}?... \textcolor{red}{christmas} is my all time} \\
      \hline
    \end{tabular}
  }
  \label{tab:example}
\end{table}

To study privacy attacks on LLMs in a cooperative learning scenario, our research begins with an exploration of information transfer in representations and progresses to a groundbreaking investigation of representation reversal attacks. 
We identify a unique challenge in inverting representations for LLMs with existing decoders: the recovery performance is significantly lower at the intermediate layer compared to the initial embedding layer in other studies~\cite{morrisLanguageModelInversion2023,morrisTextEmbeddingsReveal2023}. Embeddings contain the information of a whole sentence and pass through fewer layers compared with representations. To overcome this difficulty, we propose a two-stage inversion system with the first step involving purifying the representational information into the embedding space to solve the sparsity in representation. In the second step, we use a decoder-based generative model toward purified representation to better leverage the sequential features in our language embeddings. This research provides a robust foundation for tackling privacy and security concerns in split learning scenarios for large language models.

The contributions of this work are listed as follows:
\begin{itemize}
    \item Formal recognition and definition of severe privacy problems in the split learning scenario for LLMs.
    \item Combining information theory, this study investigates information propagation in transformer-based LLMs.
    \item Addressing the decreasing effectiveness of representation reversal after multiple layers, we propose a two-stage text recovery attack using a decoder architecture, improving state-of-the-art results by over 60\%.
\end{itemize}

\section{Related work}

In edge-cloud collaborative systems, privacy data stored on user devices may be exposed to servers or third parties through adversarial attacks. This issue is particularly critical in deep learning, where it is often referred to as the \textit{model inversion problem}. Model inversion attacks target deep learning models with the goal of inferring sensitive information from the model's output. These attacks leverage reverse engineering and reasoning techniques to reconstruct data or model parameters, posing significant privacy and security risks. Specifically, the goal of a model inversion attack is to reconstruct the training data or other sensitive information behind a given model, often through its output.

Model inversion attacks are typically classified into \textit{white-box attacks} and \textit{black-box attacks}. In a white-box scenario, the attacker has full access to the model's internal information, including its structure, weights, and outputs. In a black-box scenario, the attacker can only access the model’s output, such as confidence scores, hard labels, or feature representations. 

\subsection{Model Inversion Attacks in Computer Vision}

In the domain of computer vision, model inversion attacks typically aim to reconstruct the original image or its key features from the output of image-processing models. The concept of model inversion attacks was first introduced by Fredrikson et al\cite{fredriksonPrivacyPharmacogeneticsEndtoend2014} for simple linear regression models. Recent research has extended these attacks to more complex deep neural networks. Many computer vision models are open-source, making them prime targets for white-box attacks. These attacks can be categorized into two types: \textit{generation-based inversion} and \textit{optimization-based inversion}.

\textbf{Generation-based inversion} involves using confidence scores disclosed by the model during predictions to infer sensitive information. For example, Zhang et al.\cite{zhangSecretRevealerGenerative2020} introduced Generative Model Inversion (GMI), which leverages a generative adversarial network (GAN) to reconstruct private training data based on partial shared information. Research by Chen et al. \cite{chenkedmi} proposed KEDMI (Knowledge-Enriched Distributional Model Inversion Attacks), a new GAN-based structure that efficiently extracts private domain information from public datasets.
The black-box attack proposed by Erdogan et al. \cite{erdoganUnSplitDataObliviousModel2022} exploits the activation values transmitted through the split layers as additional knowledge for the attack.

\textbf{Optimization-based inversion} focuses on minimizing the difference between the outputs generated by a random input and the model’s output corresponding to private data. This approach iteratively adjusts the random input to converge to a reconstruction that is increasingly similar to the original data\cite{ReinforcementLearningBasedBlackBox}.

\subsection{Model Inversion Attacks in Natural Language Processing}

Similarly, language models are susceptible to privacy attacks. Yao et al. \cite{yaoSurveyLargeLanguage2023} summarize several types of privacy attacks on large language models, including adversarial attacks (e.g., data poisoning and backdoor attacks), inference attacks (e.g., attribute inference and membership inference), and extraction attacks (e.g., model stealing, gradient leakage, and training data extraction). Among these, \textit{representation inversion attacks} pose significant privacy risks as they target the hidden states of model embeddings or intermediate layers rather than requiring gradients.

\textbf{Embedding inversion attacks} are a prevalent form of representation inversion, where an attacker recovers the original text from the model's embeddings. Carlini et al. \cite{carliniExtractingTrainingData2021} showed that language models, such as GPT-2, can leak personal data (e.g., names, emails, phone numbers) through API queries. Subsequent work demonstrated that BERT \cite{devlinBERTPretrainingDeep2019} embeddings can encode significant information about the original input, enabling reconstruction of the original text.

Generative embedding inversion attacks (GEIA) were introduced by Gu et al. \cite{guSentenceLevelInference2023a}, where a generative decoder is trained to directly reconstruct the target sequence word-by-word from its embedding. Their method outperforms previous approaches, restoring semantically similar and coherent sentences from the embeddings. 

Morris et al. \cite{morrisTextEmbeddingsReveal2023} proposed Vec2Text, which introduces the idea of iteratively refining the inverted text sequence, i.e., training a self-corrector. Vec2Text first trains an embedding inversion model using the T5 model \cite{raffelExploringLimitsTransfer2023} as the base model. Then, it trains the corrector by matching the embeddings generated by the victim embedding model for the generated text with the input embeddings.

However, these approaches assume a white-box setting where the attacker has full access to the embedding model, which is often not the case in industrial settings, especially with personalized training. The assumption that embedding models' outputs can be directly targeted for inversion is insufficient for scenarios involving edge-cloud collaboration, where data representations are likely more complex and involve dynamic partitioning of models across devices and servers.

\subsection{Defense Against Model Inversion Attacks}

Despite the increasing prevalence of model inversion attacks, defense mechanisms remain underexplored. In computer vision, common defenses include adding noise to the model's output, obfuscating confidence scores, employing differential privacy, and reducing the dependency between input and output. Similar strategies are being explored in natural language processing, but adapting them to scenarios involving personalized and distributed models remains a challenge. Zhang et al. \cite{zhangSecretRevealerGenerative2020} proposed adding noise to the posterior difference during inference to thwart attacks, while Titcombe et al. \cite{titcombePracticalDefencesModel2021a} explored adding noise to representations transmitted during federated learning.

\section{Representation inversion attack}
The deployment of personalized LLMs in device-edge-cloud collaborative learning is becoming increasingly prevalent. However, privacy protection in this field remains unexplored. Inspired by intermediate representation attacks in computer vision, we suspect that there may be severe privacy leakage of input data in personalized LLM scenarios. 

\begin{figure}
\centering
\includegraphics[width=0.5\textwidth,height=0.2\textwidth]{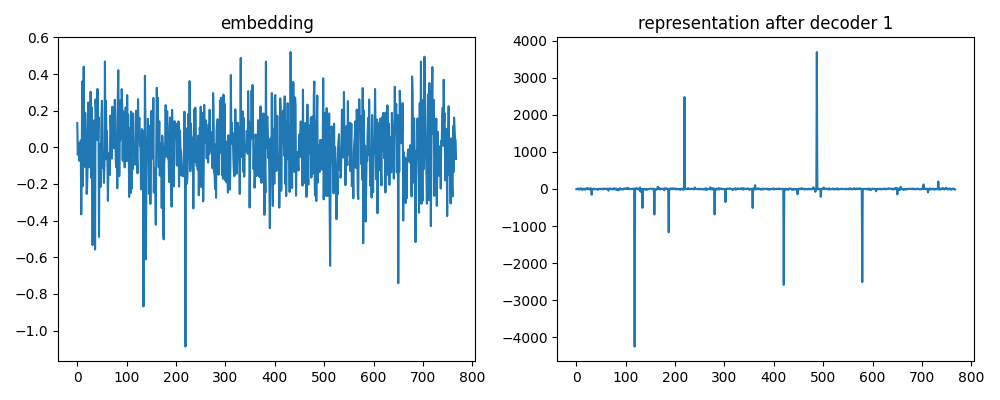}
\caption{Visualization for embedding and representation after decoder block 1} \label{analysis}
\end{figure}

As illustrated in Fig.~\ref{analysis}, the differences between embeddings and representations are significant. Post-decoding representations are considerably sparser than embeddings and contain less information about the input sentence, highlighting the need for advanced and precise attack methods.

In the following section, we formally define the problem within this new attack scenario and propose our novel attack system specifically designed to address this issue.

\subsection{Problem Formulation}


Given a sensitive input text sequence $x$ as the input of the LM, the victim large language model $g$ and personal device part of this model cut at  $i$-th layer, we denoted as $f_i$. 
Based on the autoregressive intuitive for the decoder, we define the output of $g$ at moment $t_j$ as $O_{t_j}=g(x,{O_t}_{t=0}^{t_j-1})$. 

The goal of representation inversion attacks is to reconstruct the input $x$ from its intermediate representations $\{h_i^t\}_{t=0}^{T}$, where $h_i^{t} = f_i(x,\{O_t\}_{t=0}^{t_j-1})$ and $T$ represents the final time step or maximum length of generated sequence. The top part of model $f_i$ is fixed, with its parameters and architecture not subject to modification by the adversary. Instead, the adversary leverages an auxiliary dataset $D_{\text{aux}}$ with a distribution similar to the training data to construct an external attacker model $\Phi$. This model aims to learn to reconstruct the original text, such that:
$$
\Phi(\{h_i^t\}_{t=0}^{T}) = \hat{x} \approx x.
$$

To address the computational infeasibility of enumerating all possible sequences, the problem can be approached by learning a distribution of texts given their representations. This is formulated as learning a conditional model $p(x | r; \theta)$ that maximizes the likelihood of texts given their representations:
$$
\theta = \arg\max_{\hat{\theta}} \mathbb{E}_{x\sim D}[p(x | \{h_i^t\}_{t=0}^{T}; \hat{\theta})],
$$
where $D = \{x_1, x_2, \ldots\}$ is a dataset of texts. This process essentially amortizes the combinatorial optimization of directly inverting representations into the parameters of a neural network, a task known to be challenging.

\subsection{Analysis of information}
Intuitively, after certain transformer blocks, data passes through more activation layers, leading to increased sparsity of the information. According to the On the information bottleneck theory of deep learning~\cite{shwartz-zivOpeningBlackBox2017}, as the data passes through more processors, the input information $x$ contained in the intermediate layer $h$ decreases, while the information about the output layer $y$ increases. This affects the effectiveness of model inversion attacks which motivates us to measure mutual information to study this process.

Given any two random variables, $A$ and $B$, with a joint distribution $p(a, b)$, their Mutual Information is defined as:
\begin{equation}
I(A; B) = D_{KL}[p(a, b) \| p(a)p(b)] = \sum_{a \in A, b \in B} p(a, b) \log \left( \frac{p(a, b)}{p(a)p(b)} \right)
\end{equation}
\begin{equation}
= \sum_{a \in A, b \in B} p(a, b) \log \left( \frac{p(a|b)}{p(a)} \right) = H(A) - H(A|B), \end{equation}
where $D_{KL}[p \| q]$ is the Kullback-Leibler divergence of the distributions $p$ and $q$, and $H(A)$ and $H(A|B)$ are the entropy and conditional entropy of $A$ and $B$, respectively.

\begin{figure}[h]
  \centering
  \includegraphics[width=0.5\textwidth]{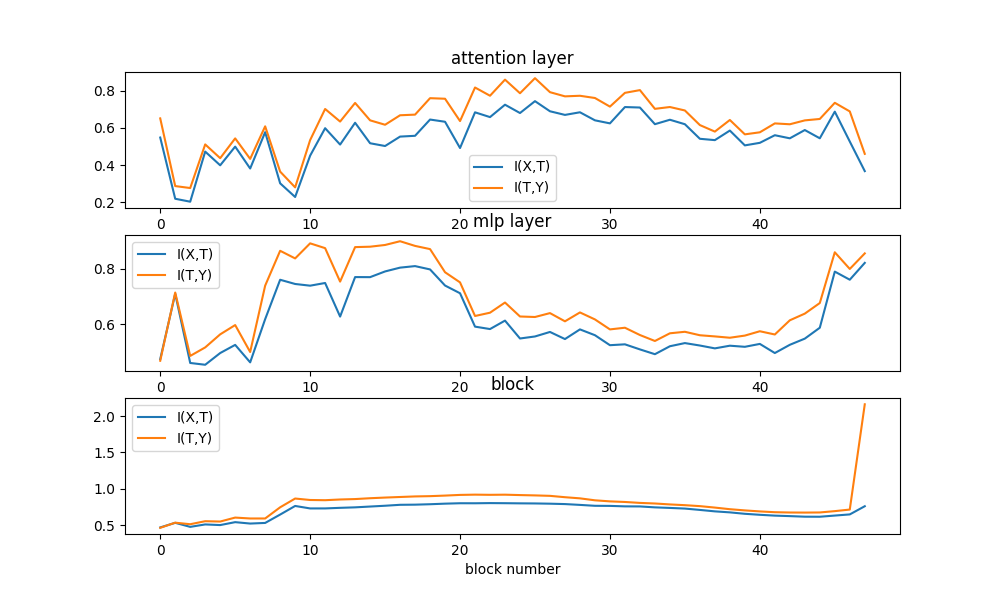}
  \caption{Mutual Information across Transformer Blocks}
\label{mi}
\end{figure}

Following \cite{shwartz-zivOpeningBlackBox2017}, for each representation $h$, we utilize discretization to compute the mutual information $I(x,h)$ and $I(h,y)$.
Fig.~\ref{mi} shows the variation of mutual information in the information propagation process for different layers $h$ across various blocks.

As the block deepens, we find that in contrast to traditional DNNs mentioned in \cite{michael2018on,shwartz-zivOpeningBlackBox2017}, the mutual information associated with the input information does not consistently decrease. Follow-up studies also found that the compression and generalization in the later stages of training mentioned in the paper are not absolute either, networks that do not compress are still capable of generalization, and vice versa. Here compression represents the reduction of $I(x,t)$ with the number of layers, which is also consistent with our findings.
This may also be related to the residual information from the specification layer added by the transformation blocks in the language model. Furthermore, we find that $I(x,t)$ and $I(t,y)$ are consistently positively correlated. In the 10th decoder block, the mutual information increases and then gradually decreases.
Inside the Transformer, we also see a rise in mutual information in the middle block. However, the amount of information passing through the forward layer is not the same as the amount of information passing through the multi-attention layer.

In summary, through the analysis of mutual information, we have discovered that the Transformer model exhibits complex dynamic variations in how it processes input and output information at different layers. This variation is influenced by various factors, including the model architecture, attention mechanism, and residual connections. Our findings enrich the understanding of the internal information flow within the Transformer, providing a theoretical foundation for the further design of attack methods.

\subsection{Threat Model}
We define the black-box attack knowledge in this scenario as follows:
The attacker has access only to the transmitted representation and the server's model structure, or the intermediate representations of each layer of the server's model.

In black-box attacks, the attacker can only use transmitted representation vectors. We propose using autoencoder-like and Transformer decoder structures for compression, due to the sparsity of representation vectors. For image data, we suggest upsampling and deconvolution networks, while for language sequences, attention mechanisms are used to better understand representation vectors.

In white-box attack, the attacker has full access to the model structure and parameters of the split large language model.
With full model knowledge, we can design more complex inversion modules, including autoencoder-like structures or those mirroring the later parts of the victim model.

\subsection{Attack method}

\begin{figure}
\includegraphics[width=0.5\textwidth]{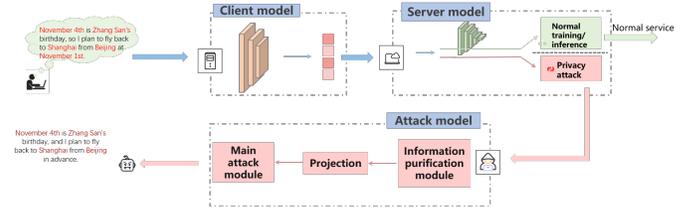}
\caption{System of RevertLM} \label{fig:llmreader}
\end{figure}

Based on the conclusion regarding the reduction of input information in intermediate layers and the fact that intermediate layer information resides in the same vector space, we propose a two-stage attack method. Specifically, due to the challenges in directly inverting deep representations, we divide the attack process into two phases. The first phase maps the representation vector to a semantically richer embedding space, referred to as the information purification module. The second phase trains an adversarial generative model as the main inversion attack module, aiming to recover the original text from the embedding vector. The attack system is illustrated in Fig. \ref{fig:llmreader}.

We then first introduces the main inversion attack module, which has strong inversion capabilities, followed by a discussion of the information purification module that addresses the first phase subproblem.


\subsubsection{Main Inversion Module} 
\label{subsec:module 2} 
Generative attack methods exhibit superior inversion and generalization capabilities compared to optimization-based approaches. Therefore, we select generative models as the primary inversion module.

Language models are structurally divided into three categories: encoder-only models, encoder-decoder models, and decoder-only models. Encoder-only models, such as BERT, are primarily used for understanding rich input semantics and excel in classification and scene comprehension tasks \cite{devlinBERTPretrainingDeep2019}. Encoder-decoder models, such as T5, are specialized in tasks that heavily depend on input, such as translation and text summarization. Decoder-only models, like the GPT series and LLaMa, are suitable for other generative tasks, including question answering. The primary inversion module is selected from encoder-decoder and decoder-only models due to their ability to leverage the generative power of large language models.

Decoder-based large language models stand out in generative tasks due to their extensive pretraining on large-scale text corpora, which enables them to capture intricate linguistic patterns. This capability allows them to generate natural language with high levels of semantic understanding, logical coherence, and fluency, excelling in both short and long-form text generation. Their robust sequence modeling ability, driven by self-attention mechanisms, extends beyond natural language processing to applications such as code generation and dialogue systems.

In this section, we utilize GPT2-XL as the attack generator. Decoder-only models, which are trained autoregressively, excel in handling complex linguistic structures and demonstrate superior generalization, making them well-suited for zero-shot and few-shot tasks. Additionally, their direct decoding of input sequences without relying on intermediate representations enables them to better capture and utilize input details, leading to more accurate and coherent outputs.

The attack module uses the representation vector as input, and the attack generator is trained using an autoregressive approach. A mapping module ensures that the input dimensions match those required by the attack generator. 





\subsubsection{Information Purification Module} 
\label{subsec:module 1} 

This section distinguishes between representation and embedding, which guides the design of the information purification module. Representation maps data to a high-dimensional space capturing its features, while embedding reduces data to a lower-dimensional space for further processing. Understanding this distinction is key to designing the purification module for various attack scenarios.

\subsubsection{Training Method} 
According to Sections~\ref{subsec:module 1} and~\ref{subsec:module 2}, the two-stage attack method in this work decomposes the complex problem by first using the information purification module to map the representation to the embedding space, and then employing the generative adversary to invert the vectors in the embedding space. 
The training method of this attack can be summarized as:

\noindent {\bf Step 1:} Pretrain information purification module with an auxiliary dataset. Direct joint training will lead to the perturbation of the generation capacity of the attack model. Our training needs an auxiliary dataset of embeddings for the victim model with negligible size. 

\noindent {\bf Step 2:} Train adversary decoder. Using the representation input after the pre-trained information purification module, we train the generative attacker with the joint loss of SequenceCrossEntropy and generative capacity loss. The SequenceCrossEntropy inspired by teacher-forcing measures 
the discrepancy between the generated token $w_i$ and the probability depends on previous tokens.
\begin{equation}
    L_{\Phi}(x; \theta_{\Phi}) = - \sum_{i=1}^{t} \log(Pr(w_i | f(x), w_0, w_1, \ldots, w_{i-1})),
\end{equation}
where $x$ is the input sequence, $w$ is the output sequence, $t$ refers to the length of output sequence.
 
We also use the Perplexity of adversary decoder as the metrics to evaluate generative capacity. The formulation is below:

$$PPL(x) = \exp\left(-\frac{1}{t} \sum_{i=1}^{t} \log Pr_{\theta_{\phi}}(w_i |  w_0, w_1, \ldots, w_{i-1})\right)$$

\noindent {\bf Step 3:} Joint fine-tune adversary decoder, projection module and autoencoder with SequenceCrossEntropy Loss. We integrate the adversary decoder, projection module, and autoencoder into a single model. This unified model is trained with a consistent optimizer.

\section{Experiments}
Our experiments include two scenarios: one involves sentiment analysis, a classification downstream task that utilizes embeddings from large language models, and the other directly uses large language models as the model for downstream tasks.

\subsection{Experiment settings}
\subsubsection{Datasets}

Most sentence embedding models are typically trained on question-answer pairs (semi-supervised tasks) and natural language inference (supervised tasks) datasets. In the experimental section of this study, we evaluate the attack performance of these models on two distinct datasets. The first selected dataset is the PersonaChat dataset \cite{zhangPersonalizingDialogueAgents2018}, which consists of open-domain conversational dialogues between two speakers with specific persona settings. Most persona settings are well represented within the corresponding dialogues, some of which may involve sensitive and private information.

The second dataset is Wiki \cite{wang-etal-2018-glue}, which is derived from the Stanford Question Answering Dataset (SQuAD) \cite{rajpurkar-etal-2016-squad} and consists of content extracted from Wikipedia articles containing domain knowledge. The presence of such domain knowledge may pose a challenge to inversion attacks. During the evaluation process, we utilize the training data of each dataset as auxiliary data to train the attacker model and report its performance on the respective test sets. A detailed summary of both datasets is provided in Table \ref{tab:dataset_llm}.

\begin{table}[!hpt]
  \caption[Statistics of Text Datasets]{Statistics of Text Datasets}
  \label{tab:dataset_llm}
  \centering
  \resizebox{0.5\textwidth}{!}{%
    \begin{tabular}{@{}lcc@{}} 
      \toprule
      \textbf{Statistic} & \textbf{PersonaChat} & \textbf{Wikipedia} \\ 
      \midrule
      \textbf{Number of Sentences} & 162,064 & 220,412 \\
      \textbf{Train/Validation/Test Split} & 82:9:9 & 95:0:5 \\
      \textbf{Number of Unique Named Entities} & 1,425 & 46,567 \\
      \textbf{Average Sentence Length} & 11.71 & 18.25 \\ 
      \bottomrule
    \end{tabular}
  }
\end{table}

\subsubsection{Metrics}
We use three standard metrics to measure the attack performance as below:
\begin{itemize}
    \item \textbf{ROUGE} represents Recall-Oriented Understudy for Gisting Evaluation \cite{linROUGEPackageAutomatic2004}. This is a metric for assessing the degree of overlap between automated abstracts or machine translations and reference summaries (or translations). It includes several variants such as ROUGE-L (scoring based on the longest common subsequence)) to measure different levels of similarity.
    \item \textbf{BLEU} score stands for Bilingual Evaluation Understudy. This is done mainly by measuring the overlap between the machine translation output and a set of reference translations.
    \item \textbf{Cosine Similarity}. This metric uses a transformer-based tokenizer and embedding model to project the two texts into a high-dimensional vector space to match similarity.
\end{itemize}

\subsubsection{Victim Models}

In our experiments, we consider different victim models and split points to evaluate the performance of the proposed inversion attack method:

\noindent \textbf{Case I}: The victim model comprises the encoder part of a language model (T5) followed by a four-layer MLP. The split point is chosen after the first layer following the embedding layer.

\noindent \textbf{Case II}: We evaluate two victim models: the decoder model GPT2-XL and the encoder-decoder model T5. The split points are strategically placed within the blocks of the decoder part or within the multi-head attention layers.

\subsection{Information in Transformer Blocks}
Through experiments summarized in Table \ref{tab:in transformer block}, we observed that the performance of the attacker model varies significantly depending on the layer from which the intermediate representations are obtained. Representations from the attention layers exhibit performance similar to those processed through entire blocks. However, representations processed by the Feed-Forward Network (FFN) layers show a significant decline in attack performance.

\begin{table}[h!]
  \centering
  \caption{Attacker performance for different layers in transformer block}
  \label{tab:in_transformer_block}
  \resizebox{0.5\textwidth}{!}{%
    \begin{tabular}{cccl}
      \toprule
      Block number & Attention blocks & FFN layer & Whole block \\
      \midrule
      Block 1 & 0.74 & 0.13 & 0.72 \\
      Block 20 & 0.66 & 0.14 & 0.62 \\
      Block 45 & 0.58 & 0.12 & 0.75 \\
      \bottomrule
    \end{tabular}
  }
\end{table}

\subsection{Attack for Different Layer Blocks}
\begin{figure}
\centering
\includegraphics[width=0.5\textwidth]{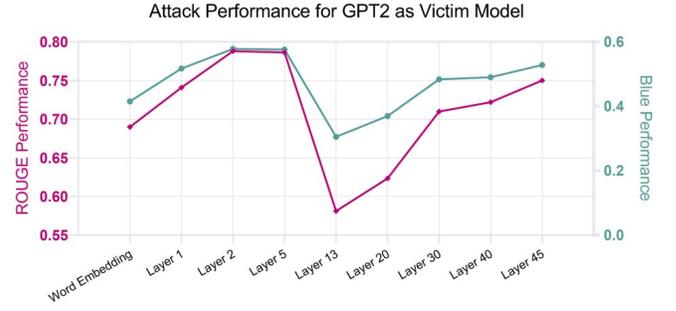}
\caption{Attack performance towards GPT2} 
\label{attack performance}
\end{figure}

Figure \ref{attack performance} illustrates the attack performance across various blocks of the GPT2 model. Initially, as the block number increases, the inversion results improve. However, there is a subsequent decline followed by an increase in performance. This trend aligns with our analysis of mutual information, indicating that intermediate representations closer to the input or output layers are more susceptible to inversion attacks.

Our base model achieves a ROUGE-L score of over 50\% and a cosine similarity recovery rate of 90\%, which are considered strong indicators of semantic similarity in NLP tasks \cite{takeshitaROUGEKYourSummaries2024,papineni-etal-2002-bleu}, regardless of the layer used as the partition point. This demonstrates that the newly proposed problem in our paper is both real and urgent. Additionally, we observe a trend where the attack effects vary depending on the segmentation layer, with some layers showing increasing vulnerability and others decreasing vulnerability to attacks.

\subsection{Different Purification Modules}
Here we formally evaluate the two-step system of RevertLM with the previously mentioned SOTA's approach GEIA \cite{guSentenceLevelInference2023a}. 

\noindent \textbf{Case I:} We evaluated the performance of the RevertLM with different purification modules. The results, presented in Table \ref{tab:different ae for 1}, indicate that the linear projection method achieves the best inversion results across all metrics. This suggests that models whose structures form an inverse function relationship with the victim model's structure can achieve better inversion results.

\begin{table}[htbp]
  \caption{Evaluation for different purification modules in RevertLM}
  \centering
  \resizebox{0.5\textwidth}{!}{%
    \begin{tabular}{cccc}
      \toprule
       & ROUGE & BLEU & Cosine Similarity \\
      \midrule
      Base RevertLM & 0.5353 & 0.2411 & 0.9209 \\
      + Linear projection & \textbf{0.5972} & \textbf{0.2985} & \textbf{0.9360} \\
      + Linear projection with tester & 0.5227 & 0.2313 & 0.9195 \\
      + Training AE & 0.5208 & 0.2279 & 0.9179 \\
      \bottomrule
    \end{tabular}
  }
  \label{tab:different_ae_for_1}
\end{table}

\noindent \textbf{Case II:} We assessed the performance of the RevertLM against the T5 model. The results, shown in Table \ref{tab:llm_personachat} and \ref{tab:llm_wiki}, demonstrate that the proposed RevertLM outperforms the state-of-the-art method across all metrics. Here T5 represents the base attack model of Vec2text~\cite{morrisTextEmbeddingsReveal2023}, note that all the models here can be used in the framework of Vec2text.

\begin{table}[!htbp]
  \centering
  \caption{Inversion Performance on the Personachat Dataset with T5 Model as the Victim Model}
  \label{tab:llm_personachat}
  \resizebox{0.5\textwidth}{!}{%
    \begin{tabular}{cccc}
      \toprule
      & ROUGE & BLEU & Cosine Similarity \\ 
      \midrule
      Vec2text & 0.1078 & 0.0004 & 0.7270 \\
      GEIA & 0.3372 & 0.0932 & 0.7841 \\
      Base RevertLM & 0.4988 & 0.2501 & 0.8663 \\
      RevertLM & 0.5630 & 0.2884 & 0.9021 \\
      \bottomrule
    \end{tabular}
  }
\end{table}

\begin{table}[!htbp]
  \centering
  \caption{Inversion Performance on the Wiki Dataset with T5 Model as the Victim Model}
  \label{tab:llm_wiki}
  \resizebox{0.5\textwidth}{!}{%
    \begin{tabular}{cccc}
      \toprule
      & ROUGE & BLEU & Cosine Similarity \\
      \midrule
      Vec2text & 0.0054 & 0.0001 & 0.8292 \\
      GEIA & 0.0504 & 0.0102 & 0.8493 \\
      Base & 0.0802 & 0.0242 & 0.8574 \\
      Ours & 0.1538 & 0.0870 & 0.8678 \\
      \bottomrule
    \end{tabular}
  }
\end{table}


\section{Conclusion}

Our study identifies critical privacy vulnerabilities in the personalized deployment of LLMs within split learning. Additionally, we analyzed the variation of mutual information across different blocks of language models, providing deeper insights into information propagation and potential leakage points. To address the challenges of attacking intermediate representations, we propose a two-step inversion system that significantly improves the recovery of these representations in split learning for large language models, enhancing recovery rates by over 60\%.

These findings underscore that intermediate representations, much like raw data, pose substantial privacy risks and must be treated with stringent anonymization and robust security measures. Protecting intermediate representations at the same level as raw text is essential to safeguard user privacy, particularly in sensitive domains like healthcare. 

\appendix

\bibliographystyle{ACM-Reference-Format}
\bibliography{main}
\end{CJK}
\end{document}